\documentclass[11pt]{article}
\usepackage{acl2016}
\usepackage{times}
\usepackage{url}
\usepackage{latexsym}
\usepackage{multirow}
\usepackage{graphicx}
\usepackage{enumitem}
\usepackage[usenames,dvipsnames]{color}
\usepackage{xspace}

\aclfinalcopy 

\newcommand{\eg}{e.g.\xspace}

\newcommand{\etc}{etc.\xspace}

\newcommand{\conf}{\texttt{CONF}\xspace}
\newcommand{\enco}{\texttt{ENCO}\xspace}
\newcommand{\ndor}{\texttt{ENDO}\xspace} 
\newcommand{\grat}{\texttt{GRAT}\xspace}
\newcommand{\fact}{\texttt{FACT}\xspace}
\newcommand{\ambi}{\texttt{AMBI}\xspace}

\newcommand\data{\textit{MedSenti}\xspace}

\newcommand\datasent{\textit{MedSenti-sent}\xspace}
\newcommand\datasubj{\textit{MedSenti-sent-subj}\xspace}

\newcommand*\rot{\rotatebox{270}}

\newcommand{\secref}[1]{Section~\ref{#1}\xspace}
\newcommand{\tabref}[2][]{Table#1~\ref{#2}\xspace}
\newcommand{\figref}[2][]{Figure#1~\ref{#2}\xspace}

\title{Toward Automatic Understanding of the Function of Affective Language in Support Groups}

\author{Amit Navindgi\thanks{\ \ Work carried out while all authors at Xerox Research Centre Europe.} \ \ \ \\
  Veritas Technologies \ \ \ \\
  Mountain View, CA \ \ \ \\
  {\tt navindgi@usc.edu} \ \ \ \\\And
  Caroline Brun, C\'ecile Boulard Masson\ \ \ \\
  Xerox Research Centre Europe \ \ \ \\
  Meylan, France \ \ \ \\  
  {\tt \{caroline.brun, \ \ \ } \\
  {\tt \ \ cecile.boulard\} \ \ \ } \\
  {\tt @xrce.xerox.com} \ \ \  \\\And
  \ \ \ Scott Nowson\footnotemark[1] \\
  \ \ \ Accenture Centre for Innovation \\
  \ \ \ Dublin, Ireland \\
  {\tt \ \ \ scott.nowson@accenture.com} \\}

\date{}

\begin{document}
\maketitle
\begin{abstract}
Understanding expressions of emotions in support forums has considerable value and
NLP methods are key to automating this. Many approaches understandably use subjective categories which are more fine-grained than a straightforward polarity-based spectrum. 
However, the definition of such categories is non-trivial and, in fact, we argue for a need to incorporate communicative elements even beyond subjectivity.
To support our position, we report experiments on a sentiment-labelled corpus of posts taken from a medical support forum. We argue that not only is a more fine-grained approach to text analysis important, but simultaneously recognising the social function behind affective expressions enable a more accurate and valuable level of understanding.

\end{abstract}



\section{Introduction}
\label{(sec:intro)}

There are a wealth of opinions on the internet. Social media has lowered the accessibility bar to an even larger audience who are now able to share their voice. However, more than just opinions on external matters, people are able to share their emotions and feelings, talking openly about very personal matters. In fact, further to merely enabling this affective expressionism, studies have shown that anonymity in online presence increases the chance of sharing more personal information and emotions when compared to face-to-face interactions~\cite{hancock2007}. 

 Medical support forums are one platform on which users generate emotion-rich content. People exchange factual information about elements such as treatments or hospitals, and provide emotional support to others with similar experiences~\cite{bringay2014}. This sharing through open discussion is known to be considerably beneficial~\cite{pennebaker2001}.

Understanding affective language in the healthcare domain is an effective application of natural language technologies. Sentiment mining on platforms such as Twitter, for example, can be seen as a quick method to gauge public opinion of government policies~\cite{speriosu2011}. However, the level of affective expression in a support forum setting is considerably more complex than a traditional positive-negative polarity spectrum.

More than just a more-fined grained labelling scheme, however, we also need a deeper understanding on the language being used. 
Much sentiment analysis research has focused on classifying the overall sentiment of a document or short text onto a positive-negative spectrum~\cite{HuL04,kim2006}. Recently, research work targeting finer grained analysis has emerged, such as aspect-based sentiment analysis \cite{2012Liu,SemEval14}, or semantic role labelling of emotions~\cite{W14-2607}. Aspect-based sentiment analysis aims at detecting fine-grained opinions expressed about different aspects of a given entity, while semantic role labelling of emotions aims at capturing not only emotions in texts but also experiencers and stimuli of these emotions. This relatively new trend in social media analytics enables the reliable detection of not simply binary sentiment, but more subtle, nuanced sentiments and mixed feelings. This is important because opinions, emotions and sentiments are not typically one-dimensional but multi-dimensional, and this is the case in any domain. Even more than this, such affective expression often serve a social purpose~\cite{rothman2016}. There is a real need to understand these specific sentiments in order to be able to pinpoint and prioritize decisions and actions to be taken according to goals and applications.

With these considerations in mind, We explore a dataset drawn from a health-related support forum, labelled for a variety of expressed sentiments. In this work we do not necessarily seek state-of-the-art performance in any specific task. We use this task to argue for two key positions:

\begin{itemize}[noitemsep]
\item that sub-document level analysis is required to best understand affective expressions
\item that to fully understand expressions of emotion in support forums, a fine-grained annotation scheme is required which takes into account the social function of such expressions.
\end{itemize}

This paper begins by reviewing work related to our propositions above. In \secref{sec:data} we describe the data which we have used, paying particular attention to the annotation scheme. We then report on our experiments, which were defined in order to support the hypotheses above. Following this, in \secref{sec:disc} we discuss the implication of this work.


 %


\section{Related Work}
\label{sec:litrev}

As reported earlier, polarity-based studies in the healthcare domain have considerable value. One work squarely in the public policy domain sought to classify tweets related to the recent health care reform in the US into positive and negative~\cite{speriosu2011}. They constructed a Twitter follower graph of relations between users, their tweets and tweets' content. They applied a semi-supervised label propagation method on unlabelled tweets and obtained better accuracy than a Maximum Entropy classifier.

\newcite{ali2013} experimented with data from multiple forums for people with hearing-loss. 
They use the subjectivity lexicon of \newcite{subjlexicon} and count-based syntactic features (\eg number of adjectives, adverbs, \etc). This approach outperformed a baseline bag-of-words model, highlighting the importance of subjective lexica for text analysis in health domain. \newcite{ofek2013} use a dynamic sentiment lexicon to improve sentiment analysis in an online community for cancer survivors. They build a domain specific lexicon from training data by representing text as bag-of-words with term-frequency as the attribute values. They train classifiers using abstract features extracted from this lexicon and outperform models trained using features extracted from a general sentiment lexicon.

\newcite{sokolova2013} took the lexicon approach further: they defined a more fine-grained annotation scheme (see \secref{sec:data} for more details) and labelled data from an IVF-related forum. Adapting Pointwise Mutual Information and using Semantic Orientation to associate n-gram phrases in the dataset with the document labels, they created a tailored, category-specific set of lexicons. These lexicons performed better, at 6-class classification, than a generic subjectivity lexicon.

In selecting their data, \newcite{sokolova2013} -- as \newcite{ali2013} and others have done -- tapped into the domain of on-line support communities. \newcite{eastin2005} showed that people who seek support on-line -- be it emotional or informational support -- typically find it. There are considerable benefits to participating in such groups. In a meta-analysis of 28 studies of health-based forums, \newcite{rains2009} reported that participants perceive an increase in social support, a significant decrease in depression, and significant increases in  both quality of life and self-efficacy in managing their condition.

Informational support is largely based around the sharing of knowledge and experiences. Emotional support is more complex, and can be framed as empathic communication. \newcite{pfeil2007} identify four components of such communication: understanding, emotions, similarities and concerns.

In addition to direct support, another common dimension of such online groups is self-disclosure~\cite{prost2012}. \newcite{barak2007} identify self-disclosure as specific to open support groups (\eg ``Cancer—Not Alone'' or ``Emotional Support for Adolescents'') as opposed to, for example, subject-specific discussion forums (\eg ``Vegetarianism and Naturalism'' or ``Harry Potter — The Book''). Self-disclosure serves three social functions \cite{tichon2003}: requesting implicit support by showing confusion and worries; providing support by sharing details of a personal experience and sharing information to further develop social relationships.

\section{Data}
\label{sec:data}

In this section we discuss aspects of the data we use in detail to support the position of the paper.

\subsection{Data Source}
The data used here\footnote{Kindly provided to us by the authors.} is that of \newcite{bobicev2015:island} -- which is an extension to the data used, and described in more depth, in \newcite{sokolova2013}. Data was collected from discussion threads on a sub-forum of an \textit{In Vitro Fertilization (IVF)} medical forum\footnote{http://ivf.ca/forums} intended for discussion by participants who belong to a specific age-group (over 35s). The dataset (henceforth \data) originally contained 1321 posts across 80 different topics. 



\subsection{Annotation Details}

There are two approaches to annotation of subjective aspects of communication: from the perspective of a reader's perception~\cite{strapparava2008} or that of the author~\cite{balahur2009}. In labelling \data \newcite{sokolova2013} opted for the reader-centric model and hence asked the annotators to analyse a post's sentiment as if they were other discussion participants. This is an important differentiation for automated classification style tasks - models are built to predict how people will understand the emotion expressed, as opposed to the emotion or sentiment an author feels they are conveying. For example, \newcite{maks2013} showed in review texts that reader assigned scores were more reliably related to the language used than the original rating of the author. Similarly, \newcite{mehl2006} showed that ratings of personality perceived by a third party ``over-hearing'' a conversation was strongly correlated with ground-truth labels acquired via self-assessment.


As discussed previously, using positive and negative labelling is too coarse for personal, empathic communications, such as those in the health support space. More fine-grained categories 
are required. The annotation scheme was evolved over multiple rounds of data exploration, with annotators consulted for their opinions on post-level sentiments as components of the evolution of thread-level sentiment. Responses were grouped and summarised and ultimately three sentiment categories were defined:

 
\begin{enumerate}[noitemsep]
\item \textbf{confusion}, (henceforth \conf) which includes aspects such as ``worry, concern, doubt, impatience, uncertainty, sadness, angriness, embarrassment, hopelessness, dissatisfaction, and dislike''
\item \textbf{encouragement}, (\enco) includes ``cheering, support, hope, happiness, enthusiasm, excitement, optimism''
\item \textbf{gratitude}, (\grat) which represents thankfulness and appreciation
\end{enumerate}


Though this set of labels was evolved independently -- to the best of our knowledge -- it captures important dimensions identified in the sociology literature. \conf here, for example, maps to expressions of confusion \cite{tichon2003} and those of concern \cite{pfeil2007}.

Broadly generalising with respect to polar labels, \conf is essentially a negative category while \enco is positive. \grat would therefore be a subset of positive expressions. In contrast, however, it was clear that certain expressions which might be considered negative on a word level -- such as those of compassion, sorrow, and pity -- were used with a positive, supportive intention. They were therefore included in the \enco category, and in fact were often posted with other phrases which would in isolation fall under this label.

In addition to the subjective categories, \newcite{sokolova2013} identified two types of objective posts: those with strictly \textit{factual} information (\fact), and those which combined factual information and short emotional expression (typically of the \enco type) which were labelled as \textit{endorsement} (\ndor). Each of the 1321 individual posts was labelled with one of the above five classes by two annotators.\footnote{Fleiss kappa $=0.73$ \cite{bobicev2015:island}.}

\subsection{Data and Label Preprocessing}

We select document labels as per \newcite{bobicev2015:island}: when two labels match, reduce to a single label; when the labels disagree the post is marked with a sixth label \textit{ambiguous} (\ambi). 

The posts in \data were taken directly from the forum. Therefore, there were a number which quoted previous posts, as is common in such discussions. For example (quoted text in \textit{italics}):

\begin{quote}
post\_id\_130007	\texttt{author1}	Member	17 September 2008 - 09:01 PM	``\textit{ \texttt{author2}, on Sep 13 2008, 10:47 AM, said:Thanks everyone.\texttt{author1} - what 3 clinics did you go to prior to Create?} Hi \texttt{author2} I went to a center outside Canada, then Lifequest and Create was the 3rd one, which I decided to go with.Hope this is helpful.\texttt{author1}''
\end{quote}

Quoted text could appear both before or after the main post content. In both cases the quote was replaced with an annotation (``\texttt{QOTEHERE}'') to indicate in future analysis that such a mechanism had been used. There were also a number of posts with quotes which contained no additional, original content. These were removed.

In addition, it is worth noting that in order to understand the relationships between document classes, we did not use the \ambi class in any experiments reported in this paper. This leaves 1137 posts in our \data corpus, with the category distribution as per \tabref{tab:cat-dist}. 




\begin{table}[ht]
\centering
\begin{tabular}{l|c|r}
\multicolumn{1}{c|}{Class} & \multicolumn{1}{|c}{\# Posts} & \multicolumn{1}{|c}{\%age} \\ 
\hline
\conf     & 115  & 10.1   \\ 
\enco & 309  & 27.2   \\ 
\ndor   & 161 &  14.2   \\ 
\grat     & 122 & 10.7    \\ 
\fact       & 430  & 37.8   \\ 
\hline
\textbf{TOTAL}         & \multicolumn{1}{|c}{\textbf{1137}} &   \\ 
\end{tabular}
\caption{Class-wise distribution}
\label{tab:cat-dist}
\end{table}

\section{Experiments}
\label{(sec:exp)}

To support our positions stated earlier regarding the understanding of affective expressions in support forums, we conducted a series of experiments. As stated, we are not seeking the achieve state-of-the-art results, but to highlight some of the challenges with current approaches.


\subsection{Broad methodology}

In this work, we use a robust dependency syntactic parser~\cite{Ait-Mokhtar2001} to extract a wide range of textual features, from word n-grams to more sophisticated
linguistic attributes. Our experiments are framed as multi-class classification tasks using liblinear~\cite{liblinear} and used 5-fold stratified cross-validation.
It is worth noting that we do not use, here, a domain-tuned lexicon. We re-implemented the Health Affect Lexicon \cite{sokolova2013} and it performed as well as previously reported. However, few studies of such approaches have shown that such lexicons generalise well, and label-based tuning is very task specific. We use the current set of categories to make more general points about work in support-related domains. 


\subsection{Document Level analysis}

In this set of experiments, we consider each post as a single unit of text with a single label. 

\subsubsection{5-class classification}

We utilised a variety of combinations of different linguistic feature sets. These ranged from basic word based n-grams, through semantic dependency features. Here, for illustrative purposes, we list the best performing combination: word uni-, bi-,  and trigrams; binary markers for questions, conjunctions and uppercase characters; and a broad-coverage polarity lexicon. Results can be seen in~\tabref{tab:5class}


\begin{table}[ht]
	\begin{center}

\begin{tabular}{l|rrr}
      & \multicolumn{1}{c}{\textit{P}} & \multicolumn{1}{c}{\textit{R}} & \multicolumn{1}{c}{\textit{F}} \\
\hline
\conf & 0.363 & 0.357 & 0.360 \\
\enco & 0.555 & 0.854 & 0.673 \\
\ndor & 0.147 & 0.062 & 0.087 \\
\grat & 0.583 & 0.492 & 0.533 \\
\fact & 0.573 & 0.502 & 0.535 \\
\hline
\textbf{MacroAvg} & 0.444 & 0.453 & 0.449 \\
\end{tabular}%

	\end{center}
	\caption{Precision, Recall and F1 for the best feature set on 5-class document-level classification}
	\label{tab:5class}
\end{table}

Our best overall score (macro averaged $F1=0.449$) is significantly above the majority class baseline ($F=0.110$). This compares favourably with the six-class performance of semantic features of the original data analysis ($F1=0.397$, Sokolova and Bobicev, 2013).
However, more important -- and not previously reported -- is the per-category performance which gives more insight into the data.  Essentially, we see that \enco, \grat and \fact perform relatively well while \conf and in particular \ndor are considerably poor.

To further explore this result we report the error matrix in \tabref{tab:5classconf}. Looking at \ndor we see that incredibly only 6\% has been correctly classified, while 86\% is classified as either \fact or \enco. This is theoretically understandable since the \ndor category is defined as containing aspects of both the other two categories directly. The reverse mis-classification is considerably less common, as is mis-classification as \grat. \conf is also mis-classified as \fact a majority, with 43\%.


\begin{table}[ht]
	\begin{center}


\begin{tabular}{l|ccccc}
   & \multicolumn{1}{c}{\rot{\conf \xspace \xspace}}   & \multicolumn{1}{c}{\rot{\enco  }} & \multicolumn{1}{c}{\rot{\ndor}} & \multicolumn{1}{c}{\rot{\grat}} & \multicolumn{1}{c}{\rot{\fact}} \\
   \hline
\conf & \textbf{36\%} & 15\%  & 3\%   & 3\%   & 43\% \\
\enco & 2\%   & \textbf{85\%} & 3\%   & 3\%   & 7\% \\
\ndor & 5\%   & 41\%  & \textbf{6\%} & 3\%   & 45\% \\
\grat & 3\%   & 30\%  & 3\%   & \textbf{49\%} & 14\% \\
\fact & 13\%  & 21\%  & 9\%   & 6\%   & \textbf{50\%} \\
\end{tabular}

	\end{center}
	\caption{Confusion Matrix -- as percentage of each class -- of best performing 5-class model.}
	\label{tab:5classconf}
\end{table}

\subsubsection{One-vs-all}

In order to further explore the pattern of errors seen in \tabref{tab:5classconf}, we attempted to directly classify each category individually in a one-vs-all binary task. The results for this task with a feature set consisting of word uni-, bi-,  and trigrams, is presented in \tabref{tab:onevall:123}. Note we report only for the target class, as the \textit{other} class is complementary.





\begin{table}[ht]
	\small
	\begin{center}

\begin{tabular}{l|rr|rrr}
      & \multicolumn{2}{|c|}{Counts} & \multicolumn{3}{|c}{Target Category}   \\
      & \multicolumn{1}{|c}{Cat} & \multicolumn{1}{c|}{Other} & \multicolumn{1}{c}{\textit{P}} & \multicolumn{1}{c}{\textit{R}} & \multicolumn{1}{c}{F}  \\
    \hline
\conf & 115   & 1022  & 0.314 & 0.070 & 0.112 \\
\enco & 309   & 828   & 0.632 & 0.728 & 0.672 \\
\ndor & 161   & 976   & 0.040 & 0.006 & 0.011 \\
\grat & 122   & 1015  & 0.697 & 0.261 & 0.365 \\
\fact & 430   & 707   & 0.642 & 0.426 & 0.492 \\
\end{tabular}%
	\end{center}
	\caption{Precision, Recall and F1 for each class in one-Vs-All setting 
    with 1-, 2-, 3-gram feature set.}
	\label{tab:onevall:123}
\end{table}

Again we see a similar pattern of performance: \enco and \fact are the most distinct from the other classes, while \conf and \ndor are considerably less so. It is worth noting that \ndor is particularly indistinct from the remainder of the corpus -- despite being the third largest category the F-score is a mere 0.011.
It is clear that this challenge is not a trivial one - there are distinct patterns of errors when classifying at the document level. In order to investigate this further, we move to sentence-level classification.


\subsection{Sentence Level analysis}

In sentence-level analysis, we tokenise each post into its constituent sentences. The 1137 posts from the \data become 8071 sentences, \datasent. Manual annotation at the sentence level is not feasible as it would be a considerably costly exercise. In order to label the corpus with the five categories of sentiment, we explored the use of automated methods.

\subsubsection{Na\"{i}ve Labelling}

The most trivial approach to label sentences is for each sentence to inherit the label of the post in which it is present. Following this na\"{i}ve method, we obtain the distribution as shown in \tabref{tab:cat-dist-sent}.
 
\begin{table}[ht]
\centering
\begin{tabular}{l|c|r}
\multicolumn{1}{c|}{Class} & \multicolumn{1}{|c}{\# Sents} & \multicolumn{1}{|c}{\%age} \\ 
\hline
\conf & 1087  & 13.5\% \\
\enco & 1456  & 18.0\% \\
\ndor & 1538  & 19.1\% \\
\grat & 733   & 9.1\% \\
\fact & 3257  & 40.4\% \\
\hline
\textbf{TOTAL}         & \multicolumn{1}{|c}{\textbf{8071}} &   \\ 
\end{tabular}
\caption{Sentence-level class distribution}
\label{tab:cat-dist-sent}
\end{table}

We run the 5-class classification scenario on \datasent using the same conditions and the previous best feature set; the results are shown in \tabref{tab:5class-sents}. Overall, the performance is worse than the post-level counterpart, with the exception of a small improvement to \ndor. \fact is the best performing individual category, though now with greater recall than precision.

\begin{table}[ht]
	\begin{center}

\begin{tabular}{l|rrr}
      & \multicolumn{1}{c}{\textit{P}} & \multicolumn{1}{c}{\textit{R}} & \multicolumn{1}{c}{\textit{F}} \\
\hline
\conf & 0.235 & 0.157 & 0.188 \\
\enco & 0.343 & 0.360 & 0.351 \\
\ndor & 0.174 & 0.088 & 0.117 \\
\grat & 0.264 & 0.225 & 0.243 \\
\fact & 0.443 & 0.598 & 0.509 \\
\hline
\textbf{MacroAvg} & 0.291 & 0.286 & 0.289 \\
\end{tabular}%

	\end{center}
	\caption{Precision, Recall and F1 for Sentence-level classification}
	\label{tab:5class-sents}
\end{table}

As with document level, we explore the model performance in more depth with the error matrix in \tabref{tab:5classsentconf}. The main observation we make here is that the drop in performance of the four subjective categories is largely due to mis-classification of sentences as \fact. Sentences in this category are the majority in \datasent (more than double the next closest category). However, the proportional differences with \data do not seem to be not enough to explain the significant changes. 

\begin{table}[ht]
	\begin{center}


\begin{tabular}{l|ccccc}
   & \multicolumn{1}{c}{\rot{\conf \xspace \xspace}}   & \multicolumn{1}{c}{\rot{\enco  }} & \multicolumn{1}{c}{\rot{\ndor}} & \multicolumn{1}{c}{\rot{\grat}} & \multicolumn{1}{c}{\rot{\fact}} \\
   \hline
\conf & \textbf{16\%} & 12\%  & 6\%   & 8\%   & 58\% \\
\enco & 6\%   & \textbf{36\%} & 12\%  & 6\%   & 41\% \\
\ndor & 7\%   & 19\%  & \textbf{9\%} & 5\%   & 60\% \\
\grat & 7\%   & 19\%  & 9\%   & \textbf{23\%} & 43\% \\
\fact & 10\%  & 14\%  & 10\%  & 6\%   & \textbf{60\%} \\
\end{tabular}

	\end{center}
	\caption{Confusion Matrix -- as percentage of each class -- of best performing 5-class model at sentence level}
	\label{tab:5classsentconf}
\end{table}

A more likely explanation is simply that the errors arise because -- at the very least -- there can be \fact-like sentences in any post. At the time of creation, annotators were asked to label ``the most dominant sentiment in the whole post'' \cite[p. 636]{sokolova2013}. For example, post 141143 contains the sentence:

\begin{quote}
Also, a nurse told me her cousin, 44, got pregnant (ivf)- the cousin lives in the USA.
\end{quote}

The post itself is labelled \enco. This sentence is -- strictly speaking -- the reporting of a fact, although it is easy to see how its purpose is to encourage others. However, at the lexico-semantic level, it is a purely objective sentence.

\subsubsection{Subjectivity-informed labelling}

One approach to re-labelling of data is to take advantage of coarser levels of annotation: that of subjectivity. Is it possible to at least distinguish which sentences are objective, and could be labelled as \fact?
We have developed a subjectivity model\footnote{citation suppressed for anonymity} built for the SemEval 2016 Aspect Based Sentiment Analysis track \cite{pontiki-EtAl:2016:SemEval}, which was among the top performing models for polarity detection. We ran the model on all sentences of the corpus in order to assess their subjectivity. Any sentence with a subjectivity likelihood of $<0.7$ (chosen  following manual assessment) we consider to be \textit{objective}; the remainder are \textit{subjective}. To attempt to eliminate the confusion with the \fact category we eliminated \textit{objective} sentences from sets previously labelled with the 4 subjective categories. In addition, because this ``confusion'' can be bi-directional, we also removed any \textit{subjective} sentences which were previously \fact. This \datasubj  set consists of 4147 sentences. Once again, we use the same experimental settings as previously, with results presented in \tabref{tab:sentences}.



\begin{table}[ht]
	\begin{center}

\begin{tabular}{l|rrr}
      & \multicolumn{1}{c}{\textit{P}} & \multicolumn{1}{c}{\textit{R}} & \multicolumn{1}{c}{\textit{F}} \\
\hline
\conf & 0.315 & 0.169 & 0.220 \\
\enco & 0.390 & 0.457 & 0.421 \\
\ndor & 0.289 & 0.126 & 0.176 \\
\grat & 0.284 & 0.294 & 0.289 \\
\fact & 0.543 & 0.745 & 0.628 \\
\hline
\textbf{MacroAvg} & 0.364 & 0.358 & 0.361 \\

\end{tabular}%

	\end{center}
	\caption{Precision, Recall and F1 for Sentence-level classification of subjectivity-adjusted corpus}		
	\label{tab:sentences}
\end{table}

Across the board, performance is marginally better with this approach (against a majority macro averaged baseline of $F1=0.107$). Importantly, in analysing the error matrix\footnote{Not presented here for space concerns.} the proportion of data mis-classified has dropped considerably (from 51\% to 37\%). However, a related consequence is that the error-rate between the \textit{subjective} categories has increased. 



\section{Discussion}
\label{sec:disc}

Despite the disappointing results in our sentence level experiments, we maintain that this level of analysis, as a step toward aspect-based understanding, is important to explore further. One reason for poor performance with both the \datasent and \datasubj is the approach to annotation at the sentence level. Naturally manual annotation of 8K sentences is considerably expensive and time consuming. However, there are clear examples in the data set of distinct labels being required. Consider the following example, (with manually annotated, illustrative labels):

\begin{quote}
post\_id\_226470 \texttt{author1} ``\textit{\texttt{author2} said [...]} $<$\texttt{ENCO}$>$ Thanks,I think we were afraid of rushing into such a big decision but now I feel it is most important not to have regrets. $<$\texttt{/ENCO}$>$ $<$\texttt{FACT}$>$ The yale biopsy is a biopsy of the lining of my uterus and it is a new test conducted by Yale University.  Here is a link you can read: URL This test is optional and I have to pay for it on my own... no coverage.$<$\texttt{/FACT}$>$''
\end{quote}

The first statement of this post is clearly intended to encourage and support the person to whom the author was responding. The second set of sentences is conveying deliberately objective, factual information about their situation. In the \data set this post is labelled as \ndor - the combination of \enco and \fact. However, the \fact component of the post is a response to a question in an even earlier post than the quoted message. It could be argued therefore that these sentiment do not relate in the way for which the \ndor label was created. To consider post-level labels, then, we would argue is too coarse grained.

To explore the possible confusion introduced by the \ndor category, particularly after removing the objective sentences in \datasubj, we conducted experiments with this category (and \fact) excluded. In this three-class experiment (\enco, \conf, and \grat), performance was again reasonable against baseline ($F1=0.510$ over $F1=0.213$), but the error rate was still high, particularly for \grat. Regardless of the linguistic feature sets, the models we have trained do not appear to be capturing the differences between the subjective categories. This seems contradictory to the original authors' intention of building ``a set of sentiments that [...] makes feasible the use of machine learning methods for automate sentiment detection.'' \cite[p. 636]{sokolova2013}. This is interesting because, from a human reader perspective (see \secref{sec:data}), the annotation scheme makes intuitive sense. That the expressions of ``negative'' emotions such as sympathy be considered in the ``positive'' category of \enco aligns with the social purpose behind such expressions \cite{pfeil2007}. Without explicitly calling attention to it, \newcite{sokolova2013} encoded social purpose into their annotation scheme. As with previous effort in the space, the scheme they have defined is very much tuned to the emotional support domain.

\begin{figure}[ht]
\includegraphics[width=\linewidth]{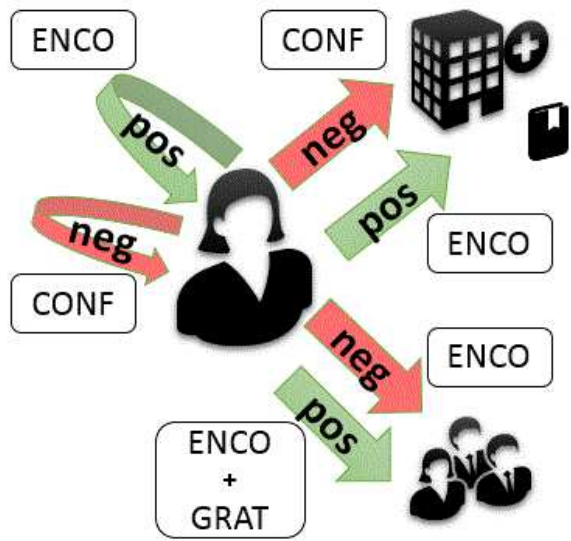}
\caption{Visualisation of polarity to category mapping given affect target -- one of either self, fellow forum participant, or external entity}
\label{fig:emoLoad}
\end{figure}

In an attempt to understand potential reasons for errors, we created a visualisation of the annotation scheme in terms of scheme category label, higher level polarity, and sentiment target, which can be seen in \figref{fig:emoLoad}.
As per the definitions of the categories, emotions expressed towards external entities, or oneself are clearly either positive-\enco or negative-\conf. However, the pattern is different in interpersonal expression between forum contributors. In the medical support environment ``negative'' expressions, as previously discussed 
serve a positive and supportive purpose. Also, the category of \grat -- a positive expression -- is always in this situation directed to another participant. This makes the interpersonal expression loadings both overloaded both in terms of classification and polarity. These relationships, in many ways, make machine modelling therein overly noisy.

Of course, it is fair to say that one direction of work in such a social domain that we did not explore is context. The original authors report subsequently on incorporating context into their experiments: both in terms of the position within a discussion of a post \cite{bobicev2015:island} and the posting history of an author \cite{sokolova2015:authors}. In this work we have eschewed context, though acknowledge that it is significantly important: in the \enco-\fact sample above, for example, context may enable a better understanding that the \enco sentence is in response to another \enco statement, while the \fact is a response to a direct question. In this sense, there is a clear motivation to understand document-level relationships at the sentence level.


Another direction which could be explored is an alternative annotation scheme.  \newcite{prost2012} suggests an annotation scheme used to identify the sharing of both practice-based and emotional support among participants of online forums for teachers. This annotation scheme is a combination of schemes developed for social support forums with those created for practice-based (\eg on-the-job, best practise discussions, or seeking of practical advice) forums. The categories identified, along with sub-categories where defined, are described in \tabref{tab:prost}.

\begin{table}[ht]
\small
\centering
\begin{tabular}{l|l}
\multicolumn{1}{c|}{\textbf{Category}} & \multicolumn{1}{c}{\textbf{Subcategory}} \\
\hline
\multicolumn{1}{l|}{\multirow{4}[0]{*}{Self disclosure}} & professional experience \\
\multicolumn{1}{l|}{} & personal experience \\
\multicolumn{1}{l|}{} & emotional expression \\
\multicolumn{1}{l|}{} & support request \\
\hline

\multirow{2}[0]{*}{Knowledge sharing} & from personal experience \\
      & Concrete info or documents \\ \hline

Opinion/evaluation & \textit{na} \\\hline

Giving advice & \textit{na} \\\hline

Giving emotional support & \textit{na} \\\hline

Requesting clarification & \textit{na} \\\hline

\multicolumn{1}{l|}{\multirow{4}[0]{*}{Community building}} & reference to community \\
\multicolumn{1}{l|}{} & humour \\
\multicolumn{1}{l|}{} & broad appreciation \\
\multicolumn{1}{l|}{} & direct thanks \\\hline

Personal attacks & \textit{na} \\
\end{tabular}
\caption{Categories and subcategories from support annotation scheme of \newcite{prost2012}}
\label{tab:prost}
\end{table}

Most of the categories are relevant for both types of forums, support and practice-based. However, the \textit{building community} category is more relevant for support forums while \textit{knowledge sharing} and and in particular \textit{personal attacks} are typically only used for practice forums.

Of course, in addition to having 15 categories, Prost annotated texts at the sub-sentence level. In order to produce the volumes of data that would be necessary for machine-learning based approaches to understanding support forum, this is impractical. There is clearly a balance to be struck between utility and practicality. However, Prost's scheme illustrates that in sociological circles, it is important to consider the social context of subjective expressions: there are two categories equivalent to \grat here, one which is more directed, and the other which concerns a bigger picture expression of the value of community.

\section{Conclusion}

In this work we have argued two positions. Despite seemingly poor results at sentence-level, we are convinced that the examples we have provided demonstrate that document-level analysis is insufficient to accurately capture expressions of sentiment in emotional support forums. We have also shown that there are important social dimensions to this type of domain which should also be taken into account. It is clear that there is considerable value to be gained from automated understanding of this increasing body of data; we in the Social NLP community need to consider some more refined approaches in order to maximise both the value itself and its fidelity.

\bibliographystyle{acl2016}


\end{document}